\PassOptionsToPackage{svgnames}{xcolor}
\PassOptionsToPackage{dvipsnames}{xcolor}
\PassOptionsToPackage{table}{xcolor}
\documentclass[acmsmall]{acmart}
\usepackage[utf8]{inputenc}
\usepackage{xcolor}
\usepackage{xspace}
\usepackage[frozencache,cachedir=.]{minted}
\usepackage{tcolorbox}
\usepackage{listings}
\setcopyright{none}
\renewcommand\footnotetextcopyrightpermission[1]{}

\usepackage{enumitem}
\setlist{topsep=0pt, leftmargin=*}

\definecolor{codegreen}{rgb}{0,0.6,0}
\definecolor{codegray}{rgb}{0.5,0.5,0.5}
\definecolor{codepurple}{rgb}{0.58,0,0.82}
\definecolor{backcolour}{rgb}{0.95,0.95,0.92}

\newcommand{\ml}{{\sc ml}}
\newcommand{\dl}{{\sc dl}}
\newcommand{\abst}{{\sc ast}}
\newcommand{\tool}{{\sc FECoM}}
\newcommand{\toolbf}{{\bf FECoM}}
\newcommand{\api}{{\sc API}}
\newcommand{\ie}{\textit{i.e.,}}
\newcommand{\eg}{\textit{e.g.,}}
\newcommand{\etal}{\textit{et al.}}
\newcommand{\viz}{\textit{viz.}}
\newcommand{\github}{{\sc GitHub}}
\newcommand{\pytorch}{{\sc PyTorch}}
\newcommand{\tf}{{\sc Tensor\-Flow}}
\newcommand{\perf}{\texttt{perf}}
\newcommand{\smi}{\texttt{nvidia-smi}}
\newcommand{\topic}[1]{\vspace{1mm}\noindent\textbf{#1:}}

\newcommand{\nbc}[3]{
		{\colorbox{#3}{\bfseries\sffamily\scriptsize\textcolor{white}{#1}}}%
		{\textcolor{#3}{\sf\small$\blacktriangleright$\textit{#2}$\blacktriangleleft$}}}
\newcommand{\todo}[1]{\nbc{TODO}{#1}{red}\xspace}

\newtcolorbox{rqbox}[1][]{
    colback=gray!10,
    colframe=gray,
    arc=1mm,
    boxrule=0.5pt,
    coltitle=black,
    fonttitle=\bfseries,
    title=#1
}

\newmintedfile{greenstyle}{
  linenos=true,
  breaklines=true,
  bgcolor=green!30
}

\newmintedfile{redstyle}{
  linenos=true,
  breaklines=true,
  bgcolor=red!30
}

\lstdefinestyle{mystyle}{
    backgroundcolor=\color{backcolour},   
    commentstyle=\color{codegreen},
    keywordstyle=\color{magenta},
    numberstyle=\tiny\color{codegray},
    stringstyle=\color{codepurple},
    basicstyle=\ttfamily\footnotesize,
    breakatwhitespace=false,         
    breaklines=true,                 
    captionpos=b,                    
    keepspaces=true,                 
    numbers=left,                    
    numbersep=5pt,                  
    showspaces=false,                
    showstringspaces=false,
    showtabs=false,                  
    tabsize=2
}

\begin{document}
\raggedbottom



\title{Enhancing Energy-Awareness in Deep Learning through Fine-Grained Energy Measurement}




\author{Saurabhsingh Rajput}
\affiliation{%
  \institution{Dalhousie University}
  \country{Canada}}
\email{saurabh@dal.ca}

\author{Tim Widmayer}
\affiliation{%
  \institution{University College London}
  \country{UK}}
\email{tim.widmayer.20@ucl.ac.uk}

\author{Ziyuan Shang}
\affiliation{%
  \institution{Nanyang Technological University}
  \country{Singapore}}
\email{zshang001@e.ntu.edu.sg}

\author{Maria Kechagia}
\affiliation{%
  \institution{University College London}
  \country{UK}}
\email{m.kechagia@ucl.ac.uk}

\author{Federica Sarro}
\affiliation{%
  \institution{University College London}
  \country{UK}}
\email{f.sarro@ucl.ac.uk}

\author{Tushar Sharma}
\affiliation{%
  \institution{Dalhousie University}
  \country{Canada}}
\email{tushar@dal.ca}

\renewcommand{\shortauthors}{Rajput, Widmayer, Shang et al.}

\begin{abstract}
With the increasing usage, scale, and complexity of Deep Learning (\dl{}) models, their rapidly growing energy consumption has become a critical concern. Promoting green development and 
energy awareness at different granularities is the need of the hour to limit carbon emissions of \dl{} systems. However, the lack of standard 
and repeatable 
tools to accurately measure and optimize energy consumption at a fine granularity (\eg{} at the \api{} level) hinders progress in this area.

This paper introduces
\toolbf{} \textbf{(\underline{F}ine-grained \underline{E}nergy \underline{Co}nsumption \underline{M}eter)}, a framework for fine-grained \dl{} energy consumption measurement. 
\tool{} enables researchers and developers 
to profile \dl{} \api{}s from energy perspective.
\tool{} addresses the challenges of fine-grained energy measurement using static instrumentation while considering factors like computational load and temperature stability.
We assess \tool{}'s capability for fine-grained energy measurement for one of the most popular open-source \dl{} frameworks, namely \tf{}. Using \tool{}, we also investigate the impact of parameter size and execution time on energy consumption, enriching our understanding of \tf{} \api{}s' energy profiles.
Furthermore, we elaborate on the considerations and challenges while designing and implementing a fine-grained energy measurement tool. 
This work will facilitate further advances in \dl{} energy measurement and the development of energy-aware practices for \dl{} systems.

\end{abstract}

\maketitle

\section{Introduction}
Deep Learning (\dl{})-based solutions are employed in an increasing number of areas concerning our day-to-day life, such as, in medicine~\cite{panesar2019machine, nayyar2021machine,bote2019deep}, transportation~\cite{veres2019deep,wang2019enhancing,haydari2020deep}, education~\cite{nafea2018machine,perrotta2020deep,vijayalakshmi2019comparison}, and finance~\cite{mehbodniya2021financial,ozbayoglu2020deep}.
However, the increasing use of \dl{} requires ample computational resources, resulting in an alarming surge in energy consumption. The extensive use of computational resources causes significantly increased {\sc co}$_2$ emissions and financial costs \cite{Fonseca19}. 
For instance, training a {\sc MegatronLM} model~\cite{Shoeybi2019}
consumes enough energy to power three American households for a year~\cite{TechTarget2021}.
This unsustainable trend is continuing; the computational resources required to train a best-in-class \ml{} model is doubling every 3.4 months~\cite{amodei_hernandez_2018}.

For this reason, it is essential to make software development, specifically using \dl{}, \textit{energy-aware}, \ie{} develop \dl{} code optimized from the energy consumption perspective, without compromising models' accuracy~\cite{SarroRE23}.
To achieve this goal, we propose to measure energy consumption for \dl{} applications at fine granularity (such as at the \api{} level) and identify energy-hungry \api{} calls; so that we can suggest alternative \textit{energy-efficient} software versions \cite{Fonseca19}.

Software engineering researchers have studied the energy footprints of software programs in recent years~\cite{desrochers2016validation,garcia2019estimation,GreenAI,10.1145/2425248.2425252}.
Broadly, energy measurement techniques are classified into \textit{hardware-based} and \textit{software-based}. 
{\it Hardware-based} techniques use physical devices such as a power monitor to measure the power consumed by a machine at a given time. 
However, hardware-based techniques are considered very difficult in use~\cite{cruz2021tools}, 
because of \textit{syncing} issues, \ie{} to sync the start and end times of the program execution with a hardware device, automatically. 
{\it Software-based} techniques measure energy by using a set of special-purpose registers, referred to as performance counters ({\sc pmc}s), in modern processors. 
These registers count specific hardware events~\cite{Khan2018}, including power consumed by hardware components, such as {\sc CPU} and memory.
Most research studies~\cite{GreenAI,abou2016trade,jurj2020environmentally} use software tools such as \perf{}~\cite{perfmanual} and \texttt{PowerStat}~\cite{canonical} that use {\sc PMC}s under the hood to measure energy consumption. 


Energy can be measured at different granularities, ranging from coarse-grained system-level to fine-grained \api{}\footnote{An \api{} (\ie{} Application Programming Interface)
refers to publicly available elements (\eg{ interfaces, classes, methods}) in a library or a framework.
An \api{} comes with its public reference documentation
that explains to the user how a method should be used, properly.
Client applications, such as \ml{}-based programs,
call these {\sc api}s from the \dl{} frameworks 
\eg{} \tf{} and \pytorch{},
to implement their functionalities.} level. 
System-level measurement considers the overall energy consumption of the entire machine or computing hardware. 
A program- or a process-level profiling examines the energy used by a software application. 
Function-level measurement profiles the energy usage within specific code blocks and methods. 
\api{}-level measurement focuses on the energy footprint of external frameworks called by the software in the form of \api{} call statements. 
\api{}-level measurement offers the finest granularity in attributing energy consumption to specific code entities.

Despite efforts to improve energy-efficiency of source code, we observe many gaps and deficiencies in the literature~\cite{Fonseca19}.
Georgiou \etal{}~\cite{GreenAI} revealed that the {\it documentation} of even the most popular \dl{} frameworks, namely \tf{} and \pytorch{}, lack an energy-consumption profile of their \api{}s.
Such an energy consumption profile could motivate software developers to explore alternative solutions and make software development more energy-efficient~\cite{Fonseca19}.
The primary reason for the lack of energy-aware documentation for \dl{} frameworks is the absence of fine-grained energy consumption measurement techniques~\cite{garcia2019estimation}.
In fact, the majority of the existing approaches allow us to measure energy consumption only at the system-level due to the support offered by hardware and operating system vendors~\cite{cruz2021tools}.

There has been some efforts to measure energy at a fine-grained level; however, existing approaches for measuring energy consumption, in general, have several deficiencies.
For example,
existing approaches~\cite{FPowerTool, javaIO}
operate at a coarser granularity, measuring entire functions and cannot be used to measure 
energy consumed by at the statement-level,
including calls to external frameworks, libraries, and APIs.
Furthermore, they support only specific programming languages such as Java, C/C++, and Fortran for CPU architectures, without considering GPU architectures used in most deep learning deployments~\cite{Dally2021}.

Software tool vendors have developed relevant tools, such as \textit{Code Carbon}~\cite{CodeCarbon} and \textit{Experiment Impact Tracker}~\cite{ExperimentImpactTracker},
to estimate power consumption and carbon emissions during the training of \dl{} models. However, these tools focus on the \ml{} \textit{program-level} granularity, leading to sampling intervals exceeding 10 seconds, which is not suitable for fine-grained energy measurements (because the measured source code entity can complete its execution in a fraction of seconds).
Moreover, they, including other academic studies so far,
overlook the overhead introduced by background processes and temperature fluctuations within the computing environment, resulting in noisy measurements.
Bannour \etal{}~\cite{bannour2021evaluating} have shown that existing tools consistently under-report energy consumption and carbon emissions, making them less sensitive to measuring energy at a smaller scale and, hence, making them unsuitable for measuring energy consumption at a finer granularity.
At present, to the best of our knowledge,
there is no convenient (\ie{} easily usable),
generic (\ie{} that can be applied on various kinds of programs and granularity),
and automated noise-free solution for measuring the energy consumption of custom deep learning code at a fine-grained level, such as at the \api{} granularity.


This study \textit{aims} to address the challenge of measuring energy consumption at the \api{} granularity as a crucial step towards using such a mechanism.
To this end, we devise a framework \viz{} \tool{}, which measures the energy consumed by \api{}s within a \dl{} framework.
\tool{} generates an Abstract Syntax Tree (\abst{}) of the input program,
applies static instrumentation by using our patching mechanism, and
measures the energy consumption of the desired \api{}s.
We empirically investigate how the size of the parameters of an \api{} call affects energy consumption and execution time (RQ2).
This empirical analysis provides valuable insights on how data size influences energy consumption,
enriching our understanding of the used \api{}s' energy profile in the context of \dl{} applications.
Though we use the developed method to measure energy consumed by a \dl{} framework \api{}s,
it can be used to measure energy consumption at a code block or even statement-level.
Given the absence of any standard for \dl{} \api{} energy-consumption at a fine-grained level of measurement, we take a step further and provide a detailed account of challenges and considerations that are vital for researchers designing energy measurement tools (RQ3). 
Such considerations will facilitate the development of more tools and techniques for energy measurement in the field.
Our study makes the following contributions:
\begin{itemize}
    \item {\bf Method and framework.} Implementing a generic method and framework, \tool{}, to accurately measure energy consumption at a fine-grained level. Such a method has been instantiated for \tf{}, to show its feasibility in practice.
    \item {\bf Static instrumentation tool.} Developing a static instrumentation tool, \textit{viz.} \textit{Patcher}~\cite{fecom_patcher_2023}, enabling necessary program patching for the energy measurement framework.
    \item {\bf Empirical study.} Conducting an empirical study to evaluate and understand the impact of parameter size on energy consumption for \dl{} \api{}s, as well as systematically examine the execution reports produced by our \tool{} to understand the reasons of failures to measure energy consumption at a fine-gained granularity.
    \item {\bf Dataset.} Create a dataset~\cite{fecom_dataset_2023} comprising energy profiling data at the \api{} granularity for $528$ \tf{} \api{} calls. This dataset covers a diverse range of domains and includes \api{} calls with varying input parameter sizes.
    \item {\bf Documenting the challenges.} Collect and categorize challenges that may arise during the development of an energy measurement tool at fine-grained granularity to facilitate the development of effective energy measurement tools and provide insights to researchers and developers in this field.
\end{itemize}
We make our framework \tool{}, the dataset, as well as the patching program used for static instrumentation publicly available~\cite{fecom_2023}.



\section{Approach} \label{section:approach}
This section describes the architecture of the \underline{F}ine-grained \underline{E}nergy \underline{Co}nsumption \underline{M}eter (\tool{}) framework and static instrumentation we devised for fine-grained energy-consumption measurement.

\subsection{\toolbf{} \space Architecture}
\begin{figure}[t]
\centering
\includegraphics[width=\linewidth]{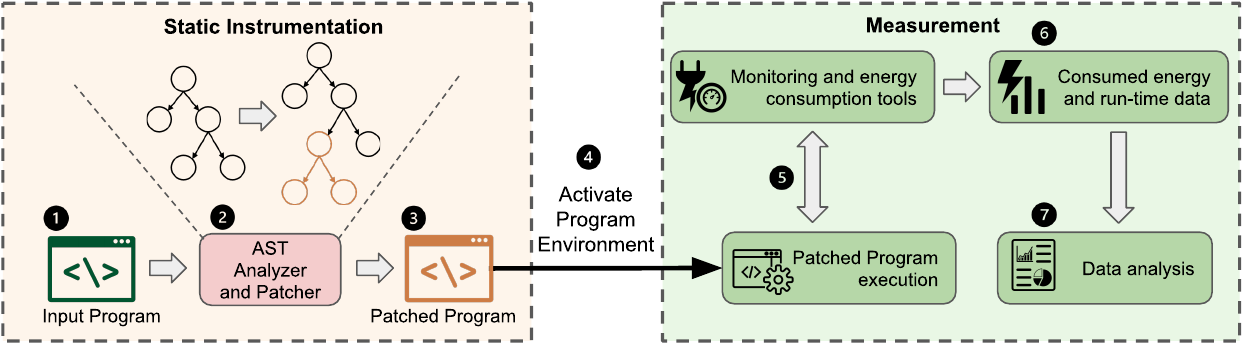}
\caption{Architecture of the \tool \space framework for energy measurement of individual methods.}
\label{fig:overview}
\end{figure}

Figure \ref{fig:overview} presents the architecture of
the proposed framework, \tool{}.
\tool{} employs static instrumentation to measure fine-grained energy consumption. 
For a given program, \tool{} identifies a set of target \api{} calls and instruments the code around the identified calls.
The instrumented code triggers \tool{}'s measurement module, enabling us to isolate the \api{} from the rest of the program during execution and measure its energy consumption.
The measurement module plays a crucial role in ensuring that the temperature and energy consumption remain stable. 
It performs the necessary checks to verify these conditions. Once the temperature and energy stability criteria are met, the measurement module proceeds to execute the \api{} call specified, capturing the corresponding energy data.
In the following, we describe each of \tool{} 's components in detail~\cite{fecom_2023}. 

\subsection{Static instrumentation}\label{static_instrumentation}
We developed a tool, referred to as \textit{Patcher}, to instrument the code to enable energy-consumption measurement.
\textit{Patcher} operates at both the method and project levels, offering fine-grained control over energy profiling.
In the context of this work, ``method'' refers to an \api{} call within the code of a \dl{} project, while ``project'' encompasses the entire codebase of a \dl{} project hosted in a repository.

\textit{Patcher} generates an Abstract Syntax Tree ({\sc ast}) of the input Python script and identifies the libraries and their aliases used within them. 
This information allows \textit{Patcher} to locate the \api{} calls corresponding to the specified libraries. 
\textit{Patcher} allows specifying the library for the analysis.
For instance, if the user wants to measure the energy consumption of the \api{}'s provided by \tf{}, 
they specify the name of the library (``\tf{}'' in this case) as an input parameter, and \textit{Patcher} automatically identifies all the \tf{} \api{} calls.
Additionally, \textit{Patcher} identifies class definitions that utilize the required libraries as base classes and keeps track of objects created from these classes. 
This enables the identification of method calls made through these objects, which are the target calls for energy measurement.

The instrumentation process inserts two source code statements around the target function calls, as shown in Listing~\ref{lst:patched_sample}, which act as breakpoints. 
The first statement, \texttt{before\_execu\-tion\_INSERTED\_\-INTO\_SCRIPT}, is placed before the original function call; the statement ensures that the machine has reached a stable state 
and captures the start time of method execution.
The second statement, \texttt{after\_execu\-tion\_INSERTED\_INTO\_SCRIPT}, is inserted after the function call; this statement records relevant information such as the total execution time and energy consumed.
We provide the list of used parameters and corresponding descriptions in Table~\ref{table:arg-description}.


The Project-level script \textit{Patcher} follows a similar approach as the method-level \textit{Patcher} by inserting the same source code statements before and after the entire script, enabling comprehensive energy-consumption measurement throughout the project's execution.



\begin{listing}[!ht]
\begin{minted}{python}
start_times_INSERTED_INTO_SCRIPT = 
before_execution_INSERTED_INTO_SCRIPT(
    experiment_file_path=EXPERIMENT_FILE_PATH,
    function_to_run="tensorflow.keras.Input()"
)
inputs = tf.keras.Input(input_shape) #Original API
after_execution_INSERTED_INTO_SCRIPT(
    start_times=start_times_INSERTED_INTO_SCRIPT,
    experiment_file_path=EXPERIMENT_FILE_PATH,
    function_to_run="tensorflow.keras.Input()",
    method_object=None,
    function_args=[input_shape],
    function_kwargs=None)
\end{minted}


\caption{Sample patched code snippet.}
\label{lst:patched_sample}
\end{listing}

\begin{table}[th]
\centering
    \caption{Description of the arguments or inserted function in the patched code.}
    \resizebox{0.75\columnwidth}{!}{%
\rowcolors{2}{gray!25}{white}
\begin{tabular}
{p{0.45\linewidth}|p{0.45\linewidth}}
\toprule
    \textbf{Argument/Function} & \textbf{Description} \\
    \midrule
    \texttt{start\_times\_INSERTED\-\_INTO\_SCRIPT} & Start times determined by before\_execution \\
    \texttt{before\_execution\_INSERTED\-\_INTO\_SCRIPT} & Patched function added before the original \api{} call \\
    \texttt{EXPERIMENT\_FILE\_PATH} & Path to store experiment data \\
    \texttt{function\_to\_run} & \api{} signature for analysis \\
    \texttt{after\_execution\_INSERT\-ED\-\_INTO\_SCRIPT} & Patched function added after the original \api{} call \\
    \texttt{method\_object} & Object in case of a method call~\eg{} model in model.compile() \\
    \texttt{function\_args} & Arguments of the \api{} call \\
    \texttt{function\_kwargs} & Keyword arguments of the \api{} call \\
    \bottomrule
\end{tabular}}
\vspace{-4mm}
    \label{table:arg-description}
\end{table}
    
\subsubsection{Validation for the static instrumentation tool}
To validate \textit{Patcher}, we drew inspiration from Automated Program Repair ({\sc APR}) techniques~\cite{wang2020automated}, adapting them to our use case. The validation process consists of the following steps: 

\topic{Executability} In the first step, we ensure the executability of the generated patches, confirming that they execute without any syntactical errors. As Python is an interpreted language, we utilize the \texttt{Pylance} language server~\cite{pylanceLanguageServer} to validate the patched code for any syntactic or type errors, ensuring a smooth compilation process.

\topic{Automated testing}
Next, we conduct automated testing on the patched scripts to verify their behavior against the original version to ensure that the patches have not introduced any unintended changes to a project (\ie{} we check if they produce the expected outputs). For this step, we randomly selected six projects from the set of projects used in our experiment (see Section~\ref{sec:dataset}); the chosen projects for evaluation represent approximately one-third of the analyzed projects.

\topic{Human evaluation}
To further validate the correctness and coverage of the code generated by \textit{Patcher}, we conducted a human evaluation.
We used the six projects selected from the previous step and sought volunteers from the Computer Science Department of Dalhousie University with prior experience developing \dl{} models in Python using \tf{}. Six graduate students were chosen to participate, and we assigned two projects to each evaluator, ensuring two evaluators for each project. The evaluators were provided with the original Python notebook, the converted Python script from the notebook, the method-level patched script, and the project-level patched script. 
They were briefed about their task using a patch template as described in Listing~\ref{lst:patched_sample}, and were informed about the purpose of the validation without indicating the authorship of \textit{Patcher}. 
The evaluators were then instructed to assess the provided artifacts and document any issues related to the following criteria:
\begin{itemize}
    \item \textbf{Correctness}: This assesses whether the generated \textit{patch} for each method adheres to the patch template as shown in Listing~\ref{lst:patched_sample}~(\ie{} it accurately extracts all the argument values for a given \api{} call). The evaluators were asked to mark each patched \api{} call as either \textit{Correct} or \textit{Incorrect} based on this criteria. To evaluate, we calculate \textit{Correctness} accuracy as the ratio of total correct patches to the total number of patches evaluated.
    \item \textbf{Completeness}: This evaluates whether the tool accurately identifies all \tf{} \api{} calls and appropriately patches all relevant calls. The evaluators were asked to provide the total number of eligible \api{} calls (\ie{}~\tf{} based calls), along with the total number of calls that were missed by the \textit{Patcher} (\ie{} \api{} calls that should have been patched, but were missed by the \textit{Patcher}). To evaluate, we calculate \textit{Completeness} accuracy as the ratio of total patched calls to the total number of eligible calls.
\end{itemize}

The evaluators were asked to fill an anonymous Excel sheet~\cite{fecom_validation_2023} for each project containing information such as logs and relevant method calls for which they checked the patched code to verify each of the evaluation criteria.
We consolidated the evaluations and checked for any differences in the evaluation for each project.


\textbf{Notably, the evaluators reported high accuracy for the patched projects based on the adopted criteria.} Specifically, the \textit{Correctness} criterion achieved $100\%$ accuracy, while the \textit{Completeness} criterion achieved $99.3\%$ accuracy. 
\textit{Patcher} missed $1$ \api{} call out of expected $159$ total calls.
A detailed analysis revealed that these missed instances are
\api{} calls made via returned \tf{} objects from user-defined functions.
The \textit{Patcher} currently 
operates on method calls made by the objects created and used in the same code block.
Hence, when a user-defined method creates and returns an object to a code block that uses the object to invoke a method, it does not get identified by the \textit{Patcher}.
We aim to address this limitation in the future versions of the \textit{Patcher}, by introducing type prediction of the returned objects.


\subsection{Pre-measurement Stability Checks}\label{stability_checks}
The machine used for the experiments must be stable before executing the \api{} calls and collecting their energy consumption as suggested by Georgiou~\etal{}~\cite{GreenAI}.
The introduced stability checks ensure low fluctuation in the hardware's energy consumption, thus reducing noise and ensuring accurate measurements by conducting energy measurement experiments under approximately the same conditions every time.

We perform two kinds of stability checks as part of the \tool{} framework---the
\textit{temperature check} and the \textit{energy stability check}.
{\sc GPU} overheating can substantially increase power draw ~\cite{price2016optimizing}, skewing results. The temperature check ensures that the {\sc CPU} and {\sc GPU} temperatures are below a standard hardware-specific threshold, maintaining uniform thermal conditions.
\tool{} uses \texttt{lm-sensors}~\cite{lm-sensors} tool to obtain the {\sc CPU} temperature and \smi{}~\cite{nvidia_developer_2022} to obtain {\sc GPU} temperature.
We follow a key guideline to run as few user processes as possible during the experiment. This guideline helps us achieve stability from a temperature and energy consumption perspective.

With energy stability check functionality, we ensure that {\sc CPU}, {\sc RAM}, and {\sc GPU} energy observations are not fluctuating. Variability indicates outside processes are consuming significant power, introducing measurement noise. The check also accounts for overheads from static instrumentation by ensuring a steady pre-instrumentation state.
Fluctuations in energy consumption indicate that other processes on the machine are consuming considerable energy, and, hence, the measured energy might include considerable noise.
To perform the check, we measure and record energy consumption by the three hardware components
(\ie{} {\sc CPU}, {\sc RAM}, and {\sc GPU}),
periodically.
We load the most recent $20$ energy data observations for the components.
Then, we determine stability by comparing the coefficient of variation ($\frac{\sigma}{\mu}$) of the data points, 
to the stable state coefficient of variation, for each component.
We calculate the stable state coefficient one time
in no-load condition \ie{}
by running only the energy measurement scripts without any other processes on the experiment machine for approximately $10$ minutes.
We calculate the coefficient of variation
by dividing the mean by the standard deviation.
If the coefficient of variation of the last $20$ observations is smaller than or equal to the stable state coefficient of variation, the machine is deemed to be in a stable state.
If both stable temperature and stable energy are achieved, the machine is ready to execute the experiments.

\subsection{Energy Measurement Module}
\label{energy-measurement-module}
The energy measurement module~\cite{fecom_measurement_2023} executes the selected project,
measures the consumed energy, and documents the observations.
We execute each project (and each \api{} call, in turn) ten times to ensure the reliability of the measurements.

\begin{figure}[t]
    \centering
    \includegraphics[width=\columnwidth]{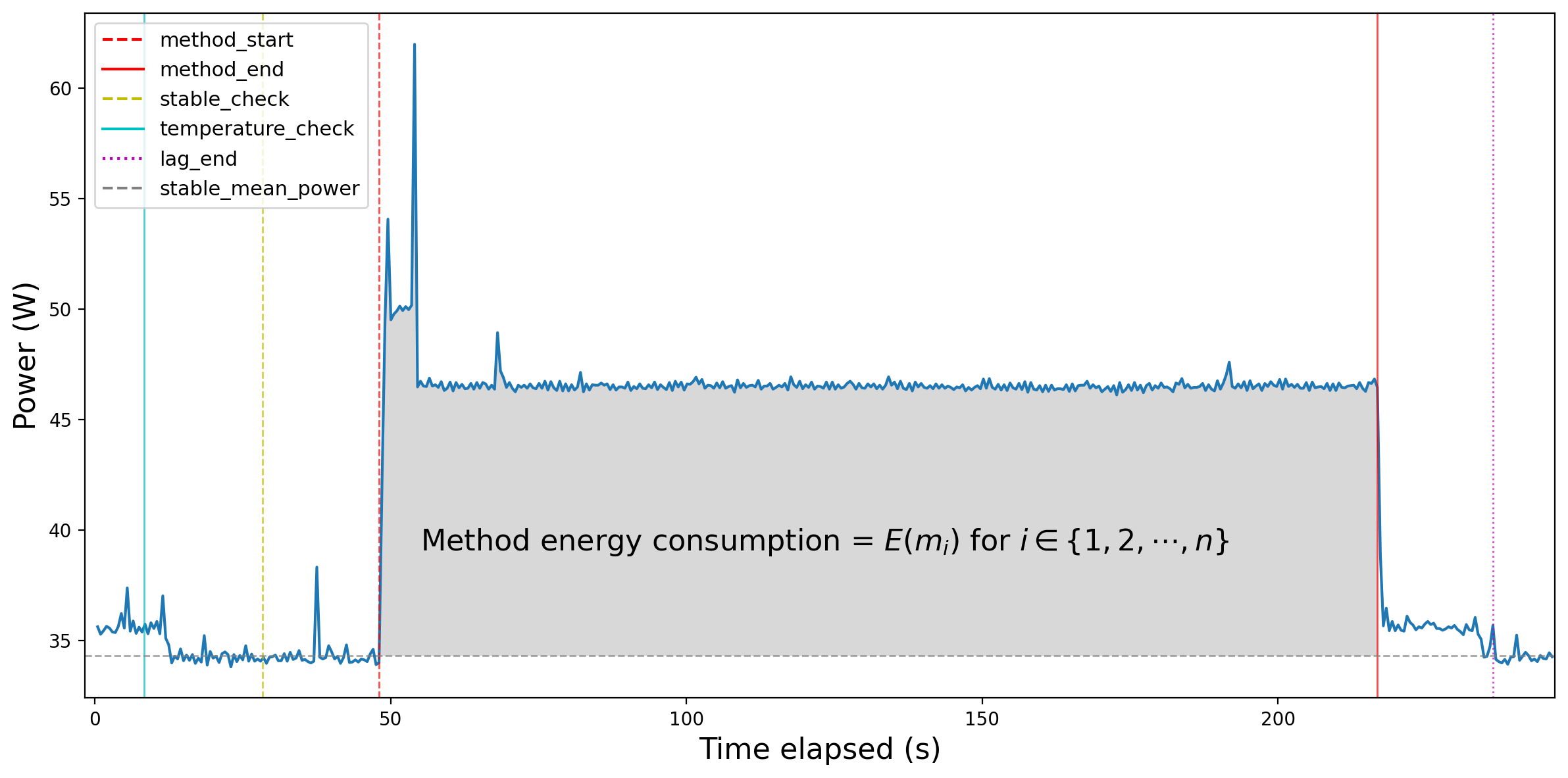}
    \caption{{\sc CPU} power over time for {\tt models.Sequential.fit} from {\tt images/cnn}.}
    \label{fig:cpu_tail_power_state}
\end{figure}

Figure~\ref{fig:cpu_tail_power_state} shows a typical power consumption profile of an \api{} call.
We measure the energy consumption by the \api{} between the start ($t_s$) and end ($t_e$) time of the call execution.
Let ${E}_t(m)$ be the energy consumed for a given \api{} call at time $t$, where $t_s \le t < t_e$.
This measurement ${E}_t(m)$ is then adjusted by subtracting the stable mean energy consumption of the system, which can be attributed to background processes, resulting in ${E}'_t(m)$. 
We measure ${E}'_t(m)$ periodically for different values of $t$ and add them to obtain $E(m)$ \ie{} $E(m) = \sum^{t_e}_{t={t_s}}{{E}'_t(m)}$.
The resulting ${E}(m)$ value is further averaged over 10 repetitions of the same experiment to obtain the mean energy consumption $\overline{E}(m)$. 
Throughout this paper, we will refer to $\overline{E}(m)$ when discussing net energy consumption in our experiments.

We store the raw measurements in JSON format, ensuring their accessibility and reproducibility via our replication package.
The collected energy consumption data include energy consumption by {\sc CPU}, {\sc RAM}, and {\sc GPU}. 
Additionally, we record timestamps, experiment settings such as wait time if the machine is unstable, wait time after \api{} call execution, stable state power consumption, {\sc GPU} and {\sc CPU} max allowed temperature, and sizes of arguments passed to the \api{} under measurement.

\subsection{Tools}
We provide below a brief description of the tools that we used for energy consumption and temperature monitoring.

    \topic{{Intel's Running Average Power Limit ({\sc RAPL})}} is an interface that allows applications to monitor and control the power consumption of various components, such as the {\sc CPU} and memory, within Intel processors. 
    {\sc RAPL} works by measuring the power consumption of the processor at regular intervals of approximately 1 ms~\cite{intel_2020} and reporting this information. It measures energy consumption of
    \begin{itemize}
        \item Package ({\sc PKG}): all {\sc CPU} components, such as cores, integrated graphics, caches and memory controllers.
        \item Core: all the {\sc CPU} cores.
        \item Uncore: all caches, integrated graphics and memory controllers.
        \item {\sc dram}: random access memory {\sc RAM} attached to the {\sc CPU}'s memory controller~\cite{dataframe_energy}.
    \end{itemize}
    Since the {\sc PKG} values include the {\sc CPU}'s total energy consumption, we will only discuss these in our analysis similar to other studies~\cite{dataframe_energy}, and we will refer to them as {\sc CPU} energy consumption.
    
\topic{Perf} is a command-line tool for collecting performance statistics from Linux systems. Data relating to energy consumption is collected using {\sc Perf}'s energy event---a wrapper around Intel's {\sc RAPL}. Specifically, the \texttt{perf stat} command is used to gather and report real-time performance counter statistics from running a command. Though \texttt{perf stat} supports reporting statistics at a maximum 1 ms frequency, 
a high overhead could result at intervals lower than 100 ms~\cite{perfmanual}.

    \topic{NVIDIA's System Management Interface (\smi{})~\cite{nvidia_developer_2022}} is a command-line utility that allows monitoring and controlling the performance of {\sc NVIDIA} {\sc GPU}s. 
    It provides detailed real-time status of the {\sc GPU}, including its power draw and temperature. 

    \topic{lm-sensors~\cite{lm-sensors}} is a Linux software tool package that enables monitoring of the hardware sensors on the {\sc CPU}, which includes temperature sensors. It provides the \texttt{sensors} command-line interface for retrieving sensor data.

\subsection{Replication Package}
The replication package for \tool{} is available on \github{}~\cite{fecom_2023}. It contains all the necessary files and instructions to replicate the experiments and results this paper presents. 

\section{Experimental Setup}
\subsection{Research Questions}
The goal of this study is to develop an approach and framework to measure energy consumption at a fine-grained level (\eg{} \api{} level) to understand better the energy profile of \api{}s of \dl{} frameworks so that it can be subsequently used to make their documentation energy-aware.
Towards this goal, we propose the framework, \tool{} (described in Section \ref{section:approach}).
We formulate the following research questions (RQs) to evaluate the proposed approach and the framework \tool{}.

\begin{enumerate}
\item [\textbf{RQ1:}] {\bf To what extent is \toolbf{} capable of measuring energy consumption at the \api{} level?} \\
  In this RQ, we aim to measure the effectiveness of the proposed framework, \tool{}. We want to ensure the correctness of the measured energy through this evaluation.
  
  \item [\textbf{RQ2:}]{\bf To what extent does input data size have an effect on energy consumption?}\\
  With RQ2, we wish to investigate the relationship between the input data provided to the \api{}s under examination and their energy consumption. 
  Answering this RQ will reveal the energy profile of \api{}s in relation to their input parameter size, which should become a critical component of energy-aware \api{} documentation.

  \item [{\bf RQ3:}] {\bf What are the main challenges and considerations in developing fine-grained energy measurement tools for \dl{} frameworks?}\\
  Verdecchia~\etal{}~\cite{verdecchia2023systematic} emphasized the significant scarcity of tools, for example, to measure energy consumption,
  in the Green Artificial Intelligence domain. Additionally, Bannour~\etal{}~\cite{bannour2021evaluating} noted that the existing tools lack sensitivity to measure energy consumption at fine granularity.
However, measuring energy consumption at lower granularities, such as \api{}-level poses unique challenges 
compared to coarse-grained measurement.
  With this research question, we aim to uncover the key challenges, underlying reasons and considerations that arise when developing tools measuring energy consumption for fine-grained profiling of \dl{} frameworks and models.
  Furthermore, answering this research question will provide insights and suggestions for researchers and practitioners, and facilitate the development of new tools and techniques in the field.
\end{enumerate}

\subsection{Experimental Design}
In this section, we elaborate on the experimental design choices and corresponding rationale for answering the research questions.

\subsubsection{\textbf{RQ1}} \label{sec:exp-design-rq1}
Validating the correctness of the measured energy consumption at a fine-grained granularity is a non-trivial challenge due to the lack of existing tools or benchmarks to measure energy consumption at the fine-grained level.
To overcome the challenge, we measure energy consumption both at the \api{} level and at the project level, \ie{} we measure the total energy consumed by the whole project and the energy consumed by all the API and method calls belonging to a framework like \tf{}.
The energy consumed by a project is approximately the sum of energy consumed by all the methods defined, called (including library/framework methods and \api{} calls), and executed within the project.
Therefore, the sum of the energy consumed by the measured methods must be less than the total energy consumed by the project.
Concretely, we model the relationship between energy consumed by a project $E(P)$ and the methods $E(m_i)$ for $i\in\{1,2,\cdots{},n\}$, in the project, in the following way.

\begin{equation}\label{eq:method_level_approx_project_level}
    E(P) \approx E(M_s) + E(M_o) = \sum^k_{i=1}{E(m_i)} + \sum^{n}_{i={k+1}}{E(m_i)}
\end{equation}

\noindent
Where methods $m_i$ for $i\in\{1,2,\cdots{},k\}$ are in the scope of energy measurement (\eg{} \tf{} methods and \api{} calls in the considered project code) representing $E(M_s)$ and, hence, measured by \tool{}.
The rest of the methods $m_i$ for $i\in\{k+1,k+2,\cdots{} n\}$
represent $E(M_o)$ that falls outside of the scope of \tool{} and, hence, we do not measure their energy consumption.
The energy consumed by methods within scope \ie{} $E(M_s)$ cannot be greater than the energy consumed by the entire project.
In Equation \ref{eq:method_level_approx_project_level}, if the energy consumed by out-of-the-scope methods is negligible, \ie{} $E(M_o) \approx 0$, we should observe $E(P) \approx E(M_s)$ if we measure the energy consumed by individual methods, correctly.

We extend the evaluation of our proposed approach by investigating
the relationship between energy consumption and execution time at the \api{} level granularity.
Previous research suggests a linear relationship between energy consumption and execution time~\cite{execution_time_vs_energy_android}, or indirectly, execution frequency~\cite{android_ml_energy_consumption}. 
This relationship seems intuitive---the longer or more frequently a task is performed, the more energy is consumed. 
This assumption holds when the power $P$ remains constant, as energy consumption $E$ is given by $E=P\times t$, where $t$ represents the time duration. 
However, if the power $P$ is a function of time $t$, the linear relationship no longer applies. 
Nevertheless, it is generally expected that as execution time increases, energy consumption will also increase proportionally.

\subsubsection{\textbf{RQ2}}\label{sec:exp-design-rq2}
The time complexity of an algorithm, $\mathcal{O}(n)$, is a function describing an asymptotic upper bound of an algorithm's run-time $t$ given an input of size $n$~\cite{sipser_theory_of_computation}. 
This implies that the execution time of an algorithm is a function of its input data size.
In Section \ref{sec:exp-design-rq1}, we discuss that energy consumption $y$ is the function of execution time \ie{} $y = f(t)$.
Therefore, energy consumption $y$ also is a function of input parameter size as given below.

\begin{equation}
    y = f(n)
\end{equation}

Though it is intuitive that the energy consumption of an \api{} call depends on the input parameter size,
the relationship between energy consumption and parameter size is not known.
This research question utilizes our proposed method and \tool{} to determine the concrete relationship between these two aspects.

In this experiment, we measure energy consumption by an \api{} call multiple times, changing the passed parameters' data size.
We determine $E_{CPU}(n)$, $E_{RAM}(n)$ and $E_{GPU}(n)$ \ie{} 
the energy consumed by {\sc CPU}, {\sc RAM}, and {\sc GPU}, respectively,
for various values of $n$.
We vary the input data size $n$ in increments of $\frac{1}{10}$ of the original data. The first run of a method uses $\frac{1}{10}$ of the original data, and this size is incremented by $\frac{1}{10}$ for each successive run until the 10$^{th}$ run uses the entire original data size.
We analyze the method-level energy consumption data from RQ1 and identify $10$ of the most energy-hungry \api{} calls.
Similar to the RQ1 setup, we execute each selected \api{} call $10$ times,
resulting in $100$ observations for each call. 

\subsubsection{\textbf{RQ3}}\label{sec:exp-design-rq3}
This RQ explores and discusses the key considerations as well as challenges that one may face while designing and developing frameworks and tools similar to \tool{} for fine-grained energy consumption measurement.
To compile a comprehensive set of design considerations and challenges, we diligently documented all the error logs and issues encountered while developing the \tool{} framework~\cite{fecom_error_2023}. 
Initially, we carefully reviewed meeting minutes from our development team comprised of the first three authors, identifying instances where roadblocks were encountered while developing the framework. 
Additionally, team members individually reviewed the initial list and augmented the identified issues.
They also added new issues based on their experience working on this study. 
We asked them to supplement the issues with additional information, including instructions to reproduce, error logs, and adopted solutions.

We used the Open Coding Technique~\cite{khandkar2009open}, a qualitative data analysis method, to analyze and categorize the identified issues. 
This approach enabled us to systematically label and categorize the collected information, identify emerging themes and patterns, and construct a conceptual framework from the raw issues. 
The process involved the following key steps:



   
 \noindent   \textbf{(1)} \textit{Coding process:} The initial step involved thoroughly reviewing the collected initial issues and deconstructing them into smaller segments for close examination. 
    We analyzed these segments to identify relationships, similarities, and dissimilarities. 
    We assigned appropriate codes to each segment to facilitate further analysis, enabling us to identify and categorize them systematically for subsequent analysis. This process was conducted in isolation for each of the team members to eliminate bias in the coding process.

 \noindent     \textbf{(2)} \textit{Iterative process:} Open coding is an iterative process that involves revisiting the segments and individual items multiple times. After an initial round of coding, we searched for new concepts and patterns by continually refining our coding scheme until we reached data saturation.
  
 \noindent     \textbf{(3)} \textit{Emergent coding:} 
    With emergent coding, codes and categories emerged directly from the data itself, allowing themes and patterns to be identified inductively. This approach prevented us from imposing preconceived notions and theories, ensuring a data-driven analysis.
  
 \noindent     \textbf{(4)} \textit{Constant comparison:} 
    We continuously compared new data with existing codes and categories throughout the coding process to ensure consistency and identify variations or similarities.

After completing the process, we obtained issues grouped into multiple categories and subcategories. 

\subsection{Repository Selection} \label{sec:dataset}
The following criteria were used to select a \dl{} project repository for evaluating the RQs.
\begin{enumerate}
    \item It should be publicly available on \github{} and popular (\ie{} having at least $5,000$ stars).
    \item It should include a good variety of \tf{}-2-based projects across various \dl{} domains and should be actively maintained.
    \item It should be easily reproducible.
\end{enumerate}

Based on these criteria, we chose the \dl{} tutorials repository from the official \tf{} documentation~\cite{tensorflow_tut} (commit: e7f81c2) for our experiments. 
This repository is publicly available on \github{}, with over $5.7$ thousand stars and $5.1$ thousand forks, and offers a wide variety of \dl{} tasks, including computer vision, natural language processing, and audio processing, each presented as a self-contained Jupyter notebook. 
These tutorials extensively utilize \tf{}'s essential \api{}'s. 
Moreover, they are designed for easy reproducibility, as each tutorial performs all the steps of the \dl{} pipeline (\ie{} data loading, pre-processing, training, testing) within the notebook without any additional dependencies. 
Additionally, the tutorials are continually updated to work with the latest \tf{} versions, maintained by a team of more than $800$ contributors.


\vspace{-2mm}
\subsection{Experimental Environment}

All experiments were conducted on a Ubuntu $22.04$ machine equipped with an Intel(R) Xeon(R) Gold 5317 {\sc CPU} (24 logical cores, $3.00$ GHz), and $125$ {\sc GB} of main memory. 
For {\sc GPU}-accelerated computations, the machine incorporates an {\sc nvidia}  {\sc GeForce rtx} $3070$ Ti with $8$ {\sc GB} GDDR6X memory. 
The {\sc GPU} exhibits an idle power of $18$ Watts and maximum power consumption of $290$ Watts.
To ensure consistent and reliable results, our experimental setup utilizes Python $3.9$ alongside \tf{} $2.11.0$. This combination requires {\sc nVIDIA} {\sc CUDA} $11.2$ and {\sc cuDNN} $8.1.0$ to leverage the full capabilities of the {\sc GPU} and optimize \dl{} computations.

\subsection{Settings}

\subsubsection{Sampling interval}\label{sec:sampling_frequency}
The frequency at which energy measurement samples are captured and retrieved is an important factor in measuring energy consumption. 
The frequency of the sampling interval affects the precision and resolution of the measurements taken. A high-frequency sampling interval can provide more precise and detailed measurements. However, it may also require more processing power and resources and generate a larger data volume (leading to more overheads)~\cite{perfmanual}. However, using a lower frequency sampling interval can also result in situations where energy consumption readings at \api{} granularity cannot be captured precisely. If a \api{}'s execution time is shorter than the sampling interval, energy consumption cannot be measured effectively. Thus, this trade-off needs consideration in selecting a sampling interval frequency.

For \tool{}, we set the sampling interval to be $500$ ms. 
The selected sampling frequency is significantly lower than the
other studies~\cite{AndroidAPIs, FPowerTool} in the field of software energy measurement.
For example,~\citeauthor{FPowerTool} uses a sampling frequency of $1$ ms.
However, \dl{} \api{}s typically have long execution times, thus making a lower frequency adequate for capturing energy fluctuations while minimizing overhead. Users of \tool{} can customize the sampling interval according to their specific experiment requirements.

\subsubsection{Machine stability}
As discussed in Section \ref{stability_checks}, maintaining the machine's temperature and energy stability is crucial to ensure accurate and reliable energy consumption measurements. We have set conservative temperature thresholds of 55$^{\circ}$C for the {\sc CPU} and 40$^{\circ}$C for the {\sc GPU}, well below the maximum allowed temperatures of 84$^{\circ}$C for the Intel(R) Xeon(R) Gold 5317 {\sc cpu}~\cite{intelXeonGold2023} and 93$^{\circ}$C for the {\sc nvidia} GeForce {\sc rtx} 3070 {\sc GPU}~\cite{nvidiaRtx3070} as specified in the product documentation.
We adopt best practices from the literature to achieve and maintain stable conditions during measurements. 
For the {\sc GPU} \smi{}, enabling persistence mode is crucial as it optimizes {\sc GPU} performance by keeping it in a fixed state and {\sc CUDA} libraries ready for immediate use, minimizing voltage and frequency changes~\cite{cai2017neuralpower, reano2015}. 
Similarly, setting the {\sc cpu} power policy to performance mode ensures it operates at maximum frequency~\cite{suchanek_navratil_domingo_bailey_2018}.

To enhance measurement accuracy, it is essential to minimize background processes on the machine related to energy measurement. Stopping unnecessary background processes~\cite{GreenAI} ensures that only the relevant processes run during the measurement cycle. Additionally, the energy measurement cycle for each \api{} call should only initiate when the machine is in a ``stable condition''~\cite{GreenAI}. 
This is ensured via the energy stability check discussed in Section \ref{stability_checks}. 
After each call execution, the machine should remain idle for a brief period to avoid tail power states~\cite{GreenAI}, ensuring the accuracy and consistency of the energy consumption measurements.

\section{Results}
In this section, we present our experimental observations for the considered research questions.

\subsection{RQ1. \toolbf{} Effectiveness}
To answer RQ1,  we measure energy consumption both at the \api{} and the project level.
As we discuss in Section~\ref{sec:exp-design-rq1}, if the sum of energy consumed by the measured \api{}s is greater than the energy consumed by the entire project, then the proposed approach is falling short of measuring energy consumption at the fine-grained granularity.
We use approximate values for the comparison because energy consumption tools have limited precision \cite{cruz2021tools, Georgiou2018}.
As we discuss in Section~\ref{sec:sampling_frequency},  we use a sampling interval of $500$ ms for the \perf{} and \smi{} tools. It implies that the energy consumption of any \api{} with a run time below $500$ ms is excluded from the sample of considered calls, further reducing the sum of method-level energy consumption values.

\begin{figure}[ht!]
    \centering
    \includegraphics[width=0.7\linewidth]{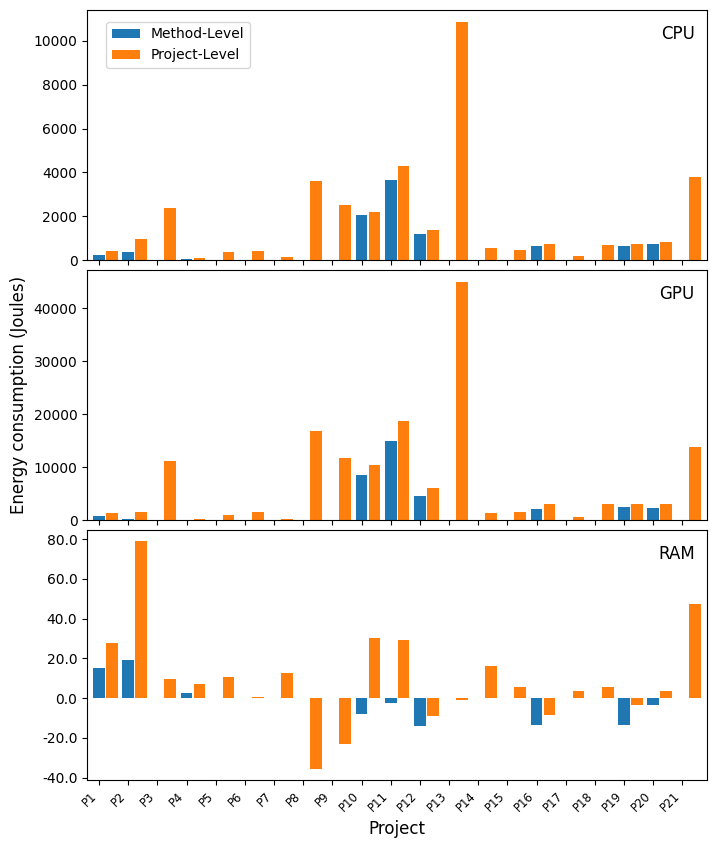}
    \caption{RQ1. Energy consumption ({\sc CPU}, {\sc GPU}, and {\sc RAM}) at project and method levels.}
    \label{fig:rq1_method_vs_project}
\end{figure}

Figure \ref{fig:rq1_method_vs_project} shows the total energy consumed by individual projects and all methods in the scope within individual projects for each hardware device (\ie{} {\sc CPU}, {\sc GPU}, and {\sc RAM}).
It shows that $E(P) > E(M_s)$ for all analyzed projects for each hardware device considered in our study. 
We can observe varying degree of difference between $E(P)$ and $E(M_s)$
for different projects. Such a variance stems from different amounts of energy consumed by methods not in the scope, \ie{} $E(M_o)$.
For instance, the project \verb|images/cnn| has $15$ \tf{} methods in scope that consume $8,513.61$ Joules energy within the {\sc GPU}; the rest of the energy $1,801.24$ Joules energy is consumed by out-of-scope $14$ methods.
Project \verb|keras/overfit_and_underfit| presents an intriguing case where we observe $E(P) >> E(M_s)$.
This difference can be attributed to the fact that the project imports \tf{} library in their source code, but it majorly utilizes a different library, other than \tf{} for \ml{} tasks.
Consequently, while the energy consumption associated with this external library is reflected in the project-level energy measurement, it remains unaccounted for in the method-level energy analysis. 
Given that \tool{} targets \tf{} methods, energy consumed by functions of other libraries is not accounted for in $E(M_s)$.

We perform statistical tests to determine the significance of the observed differences in energy consumption between method-level and project-level measurements. 
To assess the normality of the data, we conducted the Shapiro-Wilk~\cite{shapiroWilk1965} test 
at both the method and project levels with $\alpha=0.05$. The p-values for {\sc CPU} were $1.0e-04$ and $1.33e-05$, for {\sc GPU} were $3.25e-05$ and $1.80e-05$, and for {\sc RAM} were $0.001$ and $0.038$, respectively for the method and project levels. The results of the Shapiro-Wilk test indicate that the data for both the method and project levels are not normally distributed ($p-values < \alpha$).
As the data did not meet the normality assumption, we opted for non-parametric tests. 
We employed the Wilcoxon signed-rank test~\cite{wilcoxon1992individual}, which is suitable for paired samples, to assess whether the sum of method-level energy consumption is less than the project-level energy consumption for each energy type.
We performed the Wilcoxon test using the \textit{Scipy}~\cite{scipy_2023} library with the alternative hypothesis set to ``greater''  that determines whether the distribution underlying the difference between paired samples (project\_level and sum of method\_level) is stochastically greater than a distribution symmetric about zero.
Based on the results of the one-tailed Wilcoxon signed-rank test, we can conclude that there is a statistically significant difference between the method-level and project-level energy consumption for all energy types ({\sc CPU}, {\sc GPU}, and {\sc RAM}). The p-values for the Wilcoxon signed-rank tests are extremely small($<< \alpha=0.05$), with values ranging from $1.53e-05$ to $4.58e-05$, thus supporting the alternative hypothesis that the sum of method-level energy for a project is less than the project-level energy consumption.

\begin{figure}
    \centering
    \includegraphics[width=0.75\columnwidth]{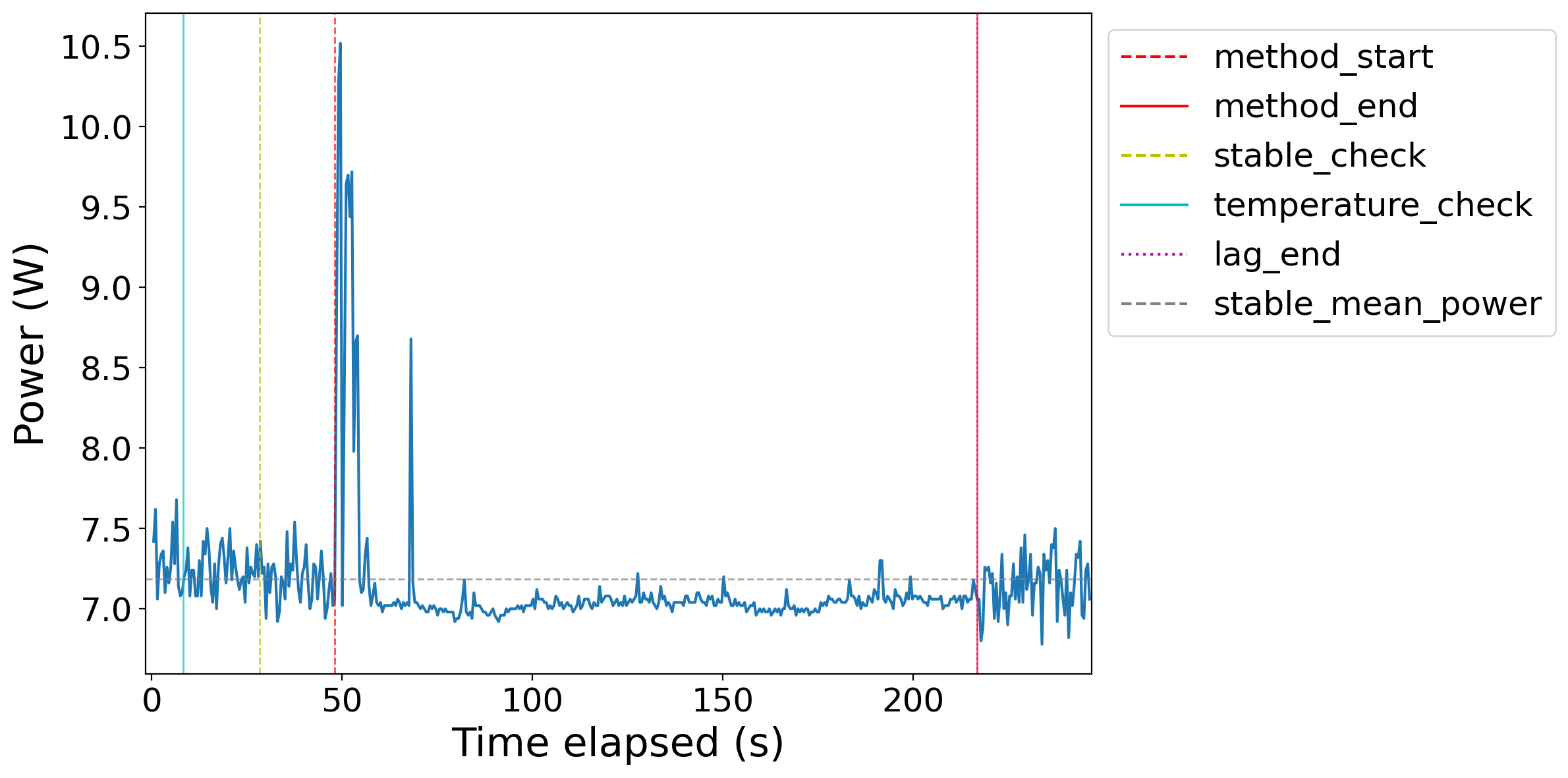}
    \caption{RQ1. {\sc RAM} power over time for {\tt models.Sequential.fit} in {\tt images/cnn}.}
    \label{fig:ram_tail_power_state}
\end{figure}

Another interesting observation stemming from Figure~\ref{fig:rq1_method_vs_project} is the presence of negative average {\sc RAM} energy consumption for some projects. 
Figure~\ref{fig:ram_tail_power_state} shows a sudden spike in {\sc RAM} power consumption at the beginning of \api{} call execution, followed by a settling down below the mean stable power.
Before executing a function on the {\sc GPU}, the input data must be copied from {\sc RAM} ({\sc CPU} memory) to {\sc GPU} memory~\cite{princeton_gpu_computing}. These extensive data copying operations could explain the spike in {\sc RAM} power draws at the start of execution.
This spike, followed by a lower {\sc RAM} power, is likely associated with \tf{} methods utilizing {\sc GPU} kernels, which utilize {\sc GPU} memory {\sc VRAM} instead of system {\sc RAM}~\cite{GpuMemoryGrowth2023}. Comparing the numerical values of $E_{CPU}(n)$ with those of $E_{GPU}(n)$ in  igure~\ref{fig:rq1_method_vs_project} confirms that {\sc GPU} energy consumption is approximately five times higher than {\sc CPU} energy consumption across all experiments. 
After the initial data copying, the {\sc CPU} and {\sc RAM} enter an idle state, requiring less energy than the stable state and, hence, the negative average energy consumption.




We extend the evaluation of our proposed approach by investigating
the relationship between energy consumption and execution time at the method-level granularity.
To validate the linear relationship between energy consumption and mean execution time, we compute the Pearson correlation coefficient. The obtained correlation coefficients $\rho$ for {\sc GPU}, {\sc CPU}, and {\sc RAM} are $0.99$ (p-value $=9.34e-27$), $0.99$ (p-value $=6.01e-39$), and $-0.84$ (p-value $=3.10e-08$), respectively. 
These values indicate that execution time for {\sc GPU} and {\sc CPU} have a strong positive relationship with energy consumption, while {\sc RAM} exhibits a strong negative relationship. 

\tool{} enables new insights into energy consumption patterns in real-world deep learning code. Consider as an example the TensorFlow program~\cite{keras_classification_tf_tutorial_2023} for image classification as shown in listing~\ref{lst:energy_sample}; this program loads data, trains a model, evaluates it, and performs inference. Using \tool{}, we can break down the total energy usage and attribute consumption to individual \api{}'s. The data loading operation via fashion\_mnist.load\_data() is quite efficient, consuming only 1J. In contrast, model.fit() for training is identified as an energy hotspot, drawing significant power at 4400J. Evaluation and inference operations exhibit more moderate consumption of 29J and 3J.

Summing the method-level measurements, the TensorFlow \api{}'s account for about 4500J energy. However, \tool{}'s project-level view shows the overall program consumes 5990J. This additional 1490J can be ascribed to non-\tf{} operations like file I/O and data pre-processing. By supporting fine-grained profiling, \tool{} reveals patterns that arise from contributions of specific API calls versus other program activities. Developers can use these insights to target optimization efforts on costly operations like fit(). \tool{} enables drilling down into energy consumption patterns within real \dl{} code, empowering developers to write greener, more efficient AI applications.


\begin{listing}[ht]
\begin{minted}[breaklines]{python}
import tensorflow as tf
....
\end{minted}
\begin{minted}[breaklines,highlightcolor=green!30,highlightlines={1-2}]{python}
# Energy efficient loading data : 1J
train_images, train_labels = fashion_mnist.load_data()  
....
\end{minted}
\begin{minted}[breaklines,highlightcolor=red!30,highlightlines={1-2}]{python}
# Energy hotspot - fit consumes maximum energy: 4400J
model.fit(train_images, train_labels, epochs=10) 
....
\end{minted}
\begin{minted}[breaklines,highlightcolor=yellow!30,highlightlines={1-2}]{python}
# Evaluation consumes moderate energy : 29J
test_loss, test_acc = model.evaluate(test_images, test_labels)
....
\end{minted}
\begin{minted}[breaklines,highlightcolor=green!30,highlightlines={1-2}]{python}
# Low energy inference: 3J
predictions = probability_model.predict(test_images)
\end{minted}
\begin{minted}[breaklines,highlightlines={2-3}]{python}
....
# Total API-level energy consumption: 4500J
# Project-level energy consumption: 5990J
\end{minted}
\caption{Example of a \tf{} code snippet annotated with energy usage}
\label{lst:energy_sample}
\end{listing}

\begin{rqbox}
    \textbf{Summary of RQ1:} The results provide strong evidence of the effectiveness of \tool{}  energy consumption measurements at a fine granularity.
\end{rqbox}



\vspace{-2mm}
\subsection{RQ2. Effect of Input Data Size on Energy Consumption}

\begin{figure}
    \centering
    \includegraphics[width=0.5\columnwidth]{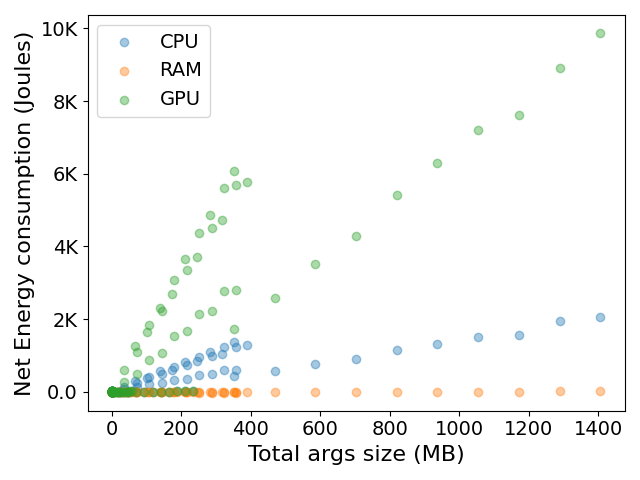}
    \caption{RQ2. Net energy consumption Vs. method call input size.}
    \label{fig:rq2_energy_vs_data_size}
\end{figure}

To answer RQ2, we executed \api{} calls with varying input data sizes. The results are presented in Figure~\ref{fig:rq2_energy_vs_data_size}. The scatter plot illustrates the Net energy consumption in Joules against the total parameter size in megabytes (MB). The figure reveals a linear relationship between input data size and energy consumption for both the {\sc CPU} and {\sc GPU}, indicating that $E_{CPU}(n)$ and $E_{GPU}(n)$ are increasing linear functions for this \api{} call. This confirms the hypothesis for RQ2. However, when considering {\sc RAM} energy consumption as a function of input data size, instead of an \textit{increasing} linear function, $E_{RAM}(n)$ appears to remain \textit{constant}.

To validate the linear relationship between energy consumption and input data size, we computed the Pearson correlation coefficient. The obtained correlation coefficients $\rho$ for {\sc GPU}, {\sc CPU}, and {\sc RAM} are $0.89$ (p-value $=4.2e-32$), $0.88$ (p-value $=3.12e-30$), and $0.19$ (p-value $=0.08$), respectively. These values indicate that input data size for {\sc GPU} and {\sc CPU} have a strong positive relationship with energy consumption, while {\sc RAM} exhibits a low positive relationship. This means that as the data size increases, {\sc CPU} and {\sc GPU} energy consumption tend to increase as well. 
The extremely small p-values for {\sc CPU}, and {\sc GPU} signify the high statistical significance of these correlations.
However, the low correlation exhibited by {\sc RAM} is not statistically significant.
The effect sizes, representing the magnitude of the Pearson correlation coefficients, are substantially large for the {\sc GPU} and {\sc CPU}, indicating a significant relationship between their energy consumption and input data size. 
As discussed in RQ1, the {\sc RAM} plays a role in the execution at the beginning of the \api{} call execution, where the data is copied to the installed {\sc GPU}. However, after 
the data is copied the role of {\sc RAM} gets over and, hence, we observe low energy consumption from the {\sc RAM} for the rest of the execution.

\begin{rqbox}
    \textbf{Summary of RQ2:}
    The results show that the energy consumption of {\sc CPU} and {\sc GPU} exhibits very strong positive correlation with the input data size. 
\end{rqbox}

\subsection{RQ3. Key considerations in Fine-grained Energy Measurement}
\begin{figure}[tb]
    \centering
    \includegraphics[width=0.8\columnwidth]{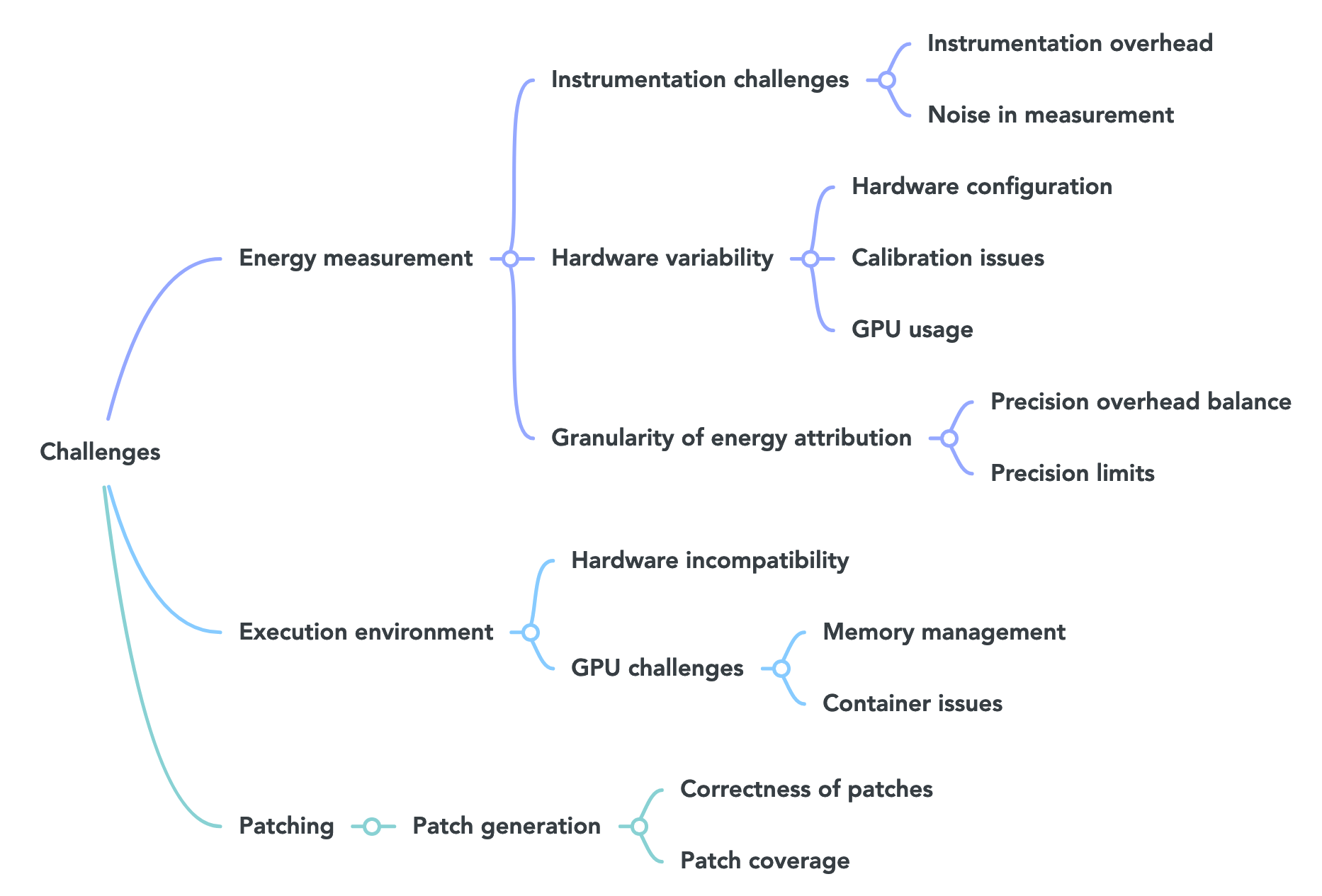}
    \caption{RQ3. Challenges in fine-grained energy measurement.}
    \label{fig:challenges}
    \vspace{-2mm}
\end{figure}

This research question aims to support developers and researchers working in this field by elaborating on the issues, considerations, and challenges one may encounter while developing a tool similar to \tool{}.
Figure \ref{fig:challenges} presents the issues and challenges (organized into categories/sub-categories) that we obtained by using an open coding process as discussed in the experiment design section~\ref{sec:exp-design-rq3}. We provide an online appendix in the replication package, with comprehensive details about the design considerations and challenges faced while developing \tool{},
along with supplementary material documenting error logs and issues used in the open coding process~\cite{fecom_error_2023}. These issues arose and were addressed inherently during \tool{}'s design and development. Prior to conducting formal experiments, we tested and refined \tool{} extensively by evaluating its functionality on a diverse set of projects. This rigorous verification was crucial for handling the complexities involved in fine-grained energy measurement.
In the rest of the section, we elaborate on each challenge
found by our analysis.

\subsubsection{\textbf{Energy measurement}}
Issues that hinder effective energy measurement
belong to this category.

\topic{Instrumentation challenges}
These challenges relate to the implementation of the energy measurement module.

\begin{itemize}
    \item \textit{Instrumentation overhead}:
        Instrumented code has additional  instructions that may account for overhead and therefore may impact the performance and energy consumption of the code.\\
        \textit{Mitigation:}
        We identify the following alternatives to address the challenge. 
        \begin{itemize}
            \item Implement machine stability and temperature stability checks.
            \item Adaptive sampling to dynamically adjust sampling frequency based on \api{} execution time.
            \item Lightweight instrumentation using binary analysis instead of source code changes.
        \end{itemize}
    \item \textit{Noise in measurement}: 
        Background processes running on the machine during energy measurement introduce noise and overheads, affecting the accuracy of measured energy. \\
        \textit{Mitigation:}
        To mitigate this issue, we suggest the following approaches.
        \begin{itemize}
            \item Ensure only necessary background processes are run during energy consumption experiments.
            \item Measure the net energy consumption by subtracting the energy of the stable baseline from the total energy consumed during measurements.
            \item Containerization or virtualization to isolate just the software being measured.
            \item Concurrent measurement on replicate systems for noise cancellation.
        \end{itemize}
\end{itemize}

\topic{Hardware variability}
Issues introduced by the heterogeneity in hardware and configurations.

\begin{itemize}
    \item \textit{Hardware configuration}: 
    Different hardware configurations can lead to variations in energy consumption values for the same project. \\
    \textit{Mitigation}: 
       To mitigate this issue, ensure all energy observations come from the same or replicated machine.
    \item \textit{Calibration issues}: 
    Energy measurement tools require calibration to account for hardware variations.\\
    \textit{Mitigation}: 
        To mitigate this issue, implement calibration processes to calculate stable energy consumption, maximum allowed temperatures, and wait times specific to the used hardware configuration that can be reused.
    \item \textit{GPU usage}: The selected subject systems must utilize {\sc GPU} in an optimized manner. Failing to ensure this challenge may introduce incorrect and inconsistent energy data collection.\\
    \textit{Mitigation}:
    To mitigate this issue, we suggest the following approaches.
    \begin{itemize}
        \item Manual inspection and execution of the program to ensure that it is indeed functional and utilizes the {\sc GPU} effectively. 
        \item Utilize monitoring tools like \texttt{NVDashboard}~\cite{nvdashboard} to monitor and visualize GPU metrics.
    \end{itemize}
\end{itemize}

\topic{Granularity of energy attribution}
Issues related to precision limit, and to the required balance between precision of energy consumption and associated overheads.
\begin{itemize}
    \item \textit{Precision limits}: 
    Existing software tools, due to Intel {\sc RAPL} limitation, do not permit energy measurements at intervals smaller than 1ms.\\
    \textit{Mitigation}:
        To mitigate this issue, take hardware \api{} support constraints into consideration while deciding the sampling rate of the profiling setup.
    \item
    \textit{Precision overhead balance}:
    Observing energy consumption at a high frequency improves the precision of observed data; however, such high frequencies also introduce computation overhead that may introduce noise.\\
    \textit{Mitigation}:
        Depending on the application context, balance the observation frequency and associated overhead.
        For instance, In our context, we measure \tf{} \api{}s that tend to run a significantly long time (minutes or, sometimes, even hours); therefore, we choose $500$ ms as the observation frequency.
\end{itemize}

\subsubsection{\textbf{Patching}} Issues related to source code instrumentation.

\topic{Patch generation}
This category summarizes the issues and considerations related to patch generation.

\begin{itemize}
    \item \textit{Correctness of patches}:
    Each identified patch location (in our case, each \tf{} \api{})
    must be correctly patched to record the correct energy consumption of the \api{} and not introduce new syntactic or semantic issues.\\
    \textit{Mitigation}:
        Ensure the correctness of the generated patches using manual validation and automated tests.
    \item \textit{Patch coverage}: 
    Each patch location must be identified correctly to avoid missing code that is supposed to be patched and measured.\\
    \textit{Mitigation}:
        Ensure the coverage of instrumentation technique by performing manual validation on the patched scripts.
\end{itemize}

\subsubsection{\textbf{Execution environment}} Environment-related issues that may hinder effective energy measurement.

\topic{Hardware incompatibility}
Compatibility issues arise when using framework versions (\eg{} \tf{}) that are not compatible with the machine's hardware or software dependencies.\\
\textit{Mitigation}:
    To mitigate the issue, focus only on subject systems that use the target framework \eg{}~\tf{}-2.
    Ensure that the required software dependencies are installed on the machine and verify the error logs to confirm that a program execution completes without any error in the logs.

\topic{GPU challenges}
Issues related to effective use of {\sc GPU} fall in this category.
    \begin{itemize}
        \item \textit{Memory management}: 
        {\sc CUDA} memory allocation errors may arise when a process cannot allocate sufficient memory during data processing on a {\sc GPU}.\\ 
        \textit{Mitigation}:
            Optimizing memory usage by managing the batch size of the training dataset and by choosing appropriate hyper-parameters can mitigate this issue.
        \item \textit{Container issues}:
        Incompatibility of Docker containers with specific {\sc GPU}s and \tf{} versions may hinder the replication of a project.\\
        \textit{Mitigation}:  
            To mitigate this issue, ensure that the NVIDIA drivers and Container Toolkit~\cite{nvidia_docker_2023} are set up on the host system. Select a suitable base image from nvidia/cuda tags~\cite{nvidia_dockerimage_2023}, and when starting containers, use the "--gpus" flag to ensure correct detection of GPUs.
    \end{itemize}



\begin{rqbox}
\textbf{Summary of RQ3:} Our experience shows that developing a tool for \dl{} \api{} fine-grained energy measurement includes several challenges, such as instrumentation overhead, noise, hardware variability, and {\sc GPU} usage. It is important to consider the granularity of energy attribution, striking a balance between precision and overheads. Correctness and coverage of generated patches are crucial for accurately recording \api{} energy consumption. Moreover, the execution environment should be carefully managed to overcome hardware incompatibility and {\sc GPU}-related challenges for effective energy measurement.
\end{rqbox}


\section{Threats to Validity}
\textbf{Internal validity:}

\noindent
\textit{Confounds and noise.} 
Several factors could affect a method's energy measurement through confounding. On the machine, multiple processes, including typical operating system processes, run in the background---scripts and tools for energy measurement, temperature checks, and stable state checks. 
All of these processes induce overheads that can skew energy consumption measurements. 
We employ several mitigation measures. 
Firstly, temperature checks are disabled before running the stable checks, and stable checks are disabled during method execution. 
This keeps the number of processes running during execution at a minimum.
Secondly, we measure stable state energy consumption without any compute load on the machine and subtract it from the gross energy consumption to get net energy consumption.

\noindent
\textit{Measurement precision.} 
Choosing an appropriate sampling interval for measuring energy at regular intervals is an important design decision.
We use a sampling interval of 500 ms.
Though considering a smaller sampling interval would have given us more observations for each experiment, it also increases the overheads and noises that may lead to incorrect energy consumption values.
Given that our target methods are framework \api{}s that typically last for minutes, if not hours, we chose a relatively low sampling interval.

\noindent
\textbf{Construct validity:}
Construct validity concerns the accuracy of the measurements and inferences using those measurements. 
To ensure accurate energy measurement, the \textit{Patcher} instruments the code. The instrumentation, coupled with stability checks, ensures that the measured energy consumption is indeed consumed by the \api{} under measurement.
We validated not only the \textit{Patcher} using automated and manual validation, but also the measured energy consumption at the \api{} granularity.

\noindent
\textbf{External validity:}
External validity deals with the generalizability of the observed results.
Given that the energy consumption is highly dependent on hardware, it may pose a threat to validity.
However, to mitigate this issue, we measure gross and net energy consumption (by reducing the gross energy consumption with stable state energy consumption).

\section{Related Work}

\noindent
{\bf Measuring energy consumption of software systems.}
A significant number of studies~\cite{sahin2014code,manotas2014seeds,hao2012estimating,sahin2012towards} uses physical power meters, 
such as the \textsc{Monsoon} high voltage power monitor~\cite{monsoon},
to measure a system's energy consumption.
These devices physically measure the electrical power consumed by a given software system.
The key benefits of using a hardware power meter are its accuracy and precision.
Another way to measure energy consumption is to use software tools~\cite{vsimunic1999cycle,simunic2000source,brooks2000wattch,gurumurthi2002using}.
Recent Intel and {\sc amd} processors provide the \textit{Running Average Power Limit} ({\sc rapl}) interface~\cite{weaver}, which can measure the power consumption of a processor at regular intervals
through built-in performance counters.
{\sc rapl} can measure the power consumption up to intervals of approximately 1 ms, translating to a sampling frequency of 1 kHz~\cite{intel_2020}.
Many tools have been built on top of {\sc rapl}, such as the Intel {\sc Power Gadget} (it has been discontinued from usage)~\cite{intel_2019}, {\sc PowerTOP}~\cite{powerTOP} and {\sc Perf}~\cite{perfmanual}.
Power modeling is another method that is used to obtain insights into the energy consumption of software systems~\cite{hao2012estimating}.
Power modeling techniques estimate energy consumption by considering factors such as the energy characteristics of the hardware and run-time information.
However, power modeling techniques' full potential has yet to be unlocked.

\begin{table}[th]
\centering
\caption{Comparison of energy measurement techniques}
\label{tab:rw-comparison}
\resizebox{0.98\columnwidth}{!}{
\rowcolors{2}{gray!25}{white}
\begin{tabular}{p{0.18\linewidth}|p{0.18\linewidth}|p{0.14\linewidth}|p{0.14\linewidth}|p{0.24\linewidth}|p{0.14\linewidth}|p{0.14\linewidth}|p{0.14\linewidth}|p{0.14\linewidth}}
\toprule
\textbf{Approach} & \textbf{Granularity} & \textbf{Sampling Rate} & \textbf{Languages} & \textbf{Architecture} & \textbf{DL Frameworks} & \textbf{Power stability} & \textbf{Temperature stability}  & \textbf{Automated}\\
\midrule

CodeCarbon~\cite{CodeCarbon} & Program/Function  & 15s & Python & \textsc{CPU(Deprecated), GPU and RAM} & $\checkmark$ & $\times$& $\times$& $\times$\\

FPowerTool~\cite{FPowerTool} & Function & 1ms & Fortran/C/C++ & \textsc{CPU}& $\times$ & $\times$& $\times$& $\times$\\

JavaIO~\cite{javaIO} & System & 1ms & Java & \textsc{CPU}& $\times$& $\times$ & $\times$& $\times$\\

Perf~\cite{perfmanual} & System & 1ms & n/a & \textsc{CPU}& $\times$& $\times$ & $\times$& $\times$\\

PowerTOP~\cite{powerTOP} & System & 20s & n/a & \textsc{CPU and GPU} & $\times$& $\times$ & $\times$& $\times$\\

Experiment Impact Tracker~\cite{ExperimentImpactTracker} & Program/Function & 1s & Python & \textsc{CPU and GPU}& \checkmark& $\times$& $\times$& $\times$\\

\textbf{FECoM} & \textbf{\textsc{API}/Program} & \textbf{500ms} & \textbf{Python} & \textbf{\textsc{CPU, GPU and RAM}} & \textbf{\checkmark} & \checkmark & \textbf{\checkmark}& \textbf{\checkmark}\\

\bottomrule
\end{tabular}
}
\end{table}


\noindent
{\bf Optimizing energy consumption of ML tasks.}
Previous studies~\cite{GreenAI, zhang2018pcamp, anthony2020carbontracker, henderson2020towards} have investigated the energy consumption of different \dl{} models. 
Algorithmic optimization is
a major approach that contributes to reducing the energy consumption of \dl{} models.
These optimizations often take place in model pruning,
which refers to removing unnecessary connections within a neural network~\cite{Song2016, Yang_2017_CVPR}.
and quantization, which reduces the precision of the weights and activations within a neural network by making them quantized and reducing the memory required for a given model.
For instance, Han et al.~\cite{Song2016} found that the energy consumption of deep neural networks can be reduced by using a combined technique consisting of model pruning, weight quantization, and Huffman coding.
This approach can reduce the size and computational complexity of the deep learning models without significant degradation to their performance.

\noindent
{\bf Measuring energy consumption at different granularities.}
Available hardware and software tools measure energy consumption at the system level.
To improve the granularity of analysis and recommendations,
researchers have attempted 
to measure the energy consumption at a more fine-grained level, such as at the \textit{process} level~\cite{AndroidAPIs, javaIO} at best.
Bree et al.~\cite{bree2022energy} attempted method replacement for measuring the energy consumed by the visitor pattern; however, their approach is unsuitable for measuring any given method’s energy consumption due to the lack of a generalizable solution. FPowerTool~\cite{FPowerTool} 
operates at a coarser granularity of function blocks for programming languages written in C, C++, and Fortran in non-dl{} programs. However, its biggest drawback is that it uses a dynamic instrumentation technique that involves injecting tracing code during the runtime, which would lead to noisy and inaccurate energy measurement data at finer granularities such as \api{} level.
Similarly, \textsc{CodeCarbon}~\cite{CodeCarbon} and \textsc{Experiment Impact Tracker}~\cite{ExperimentImpactTracker} are open-source software tools designed to monitor and
reduce the {\sc co}$_2$ emissions associated with computing processes, particularly those involved in \dl{} applications.
These tools integrate themselves into Python codebases and enable the tracking of emissions based on power consumption and location-dependent carbon intensity.
While CodeCarbon and Experiment Impact Tracker can measure power consumption, and hence {\sc co}$_2$ emissions at a function-level granularity through the use of decorators, they have a few drawbacks.
Firstly, the process of modifying source code to measure the power consumption for desired methods is manual in nature. For instance, if users of CodeCarbon wish to measure the power consumption of \tf{} \api{} calls, they have to insert the addition lines-of-code calling the CodeCarbon tool themselves, manually for each \tf{} \api{} instance.
Secondly, and more importantly, it overlooks the overheads introduced by background processes and temperature fluctuations while having a sampling rate of $15$ seconds, leading to significant noise in the measured energy consumption.
Therefore, to the best of our knowledge,
the literature does not offer any approach that measures energy consumption
at the method level for \dl{} frameworks.
{\bf Gaps in existing research.}
The existing literature as shown in Table~\ref{tab:rw-comparison} has two critical limitations.
First, existing approaches can measure energy consumption at only the system level due to support offered by hardware and operating system vendors. 
There have been efforts to measure energy consumption at the finer granularity (such as at the function level); however, as discussed, they lack various requirements like low sampling rate, stability check, support for \dl{} framework and languages like \tf{} and python, as well as support for {\textsc{CPU, GPU and RAM}} architectures.
Second, current approaches require manually instrumenting the code to measure energy consumption. 
For a software engineer to improve the energy efficiency of a given code conveniently and efficiently, the engineer must have access to an approach that can automate the energy measurement process to work at the fine-grained level so that the engineer can take corrective actions, if needed, early. 
Our proposed approach addresses the gap by providing an automated and generic mechanism to measure energy consumption at the method granularity.

\label{rw}

\section{Implications}

\subsection{Implications for Researchers}

Researchers in the field can build upon the foundation of \tool{} to develop hybrid measurement approaches, especially using the findings from RQ3. Researchers can conduct empirical studies to gain insights into energy consumption patterns, revealing relationships between model architecture, hyperparameters, and energy efficiency, contributing to the development of more energy-efficient \dl{} models and algorithms. Researchers could leverage \tool{} to construct detailed energy profiles of DL models, illuminating optimization opportunities. For example, in RQ2, we showed how \tool{} can be used to gain insights into the relationship between energy consumption and input data size.
These insights are steps towards enriching \api{} documentation of \dl{} frameworks. Through extensions and new experiments with \tool{}, researchers can gain a deeper understanding of energy dynamics in DL systems.
\subsection{Implications for Developers}
DL developers can utilize \tool{}'s capabilities to measure and optimize energy consumption at the granularity of framework APIs. Developers can pinpoint energy hotspots in their code down to the API call level by incorporating \tool{} into their development workflows. For example, in RQ1, we observed that for the Autoencoder project~\cite{autoencoder_tf_tutorial_2023}, the same type of API call ``Model.fit'' consumed varying amounts of energy ($5656.93$ J, $7330.60$ J, $133.20$ J) based on the context of the call. This fine-grained profiling allows developers to make informed optimizations like substituting inefficient APIs, streamlining data pipelines, and adopting energy-aware coding practices. \tool{} ultimately enables developers to build greener, leaner DL applications by illuminating energy consumption patterns. Its ease of use and integration with popular frameworks like TensorFlow can encourage energy awareness among the wider DL developer community.
\subsection{Implications for Educators} 
As DL courses expand, educators can utilize \tool{} to instill energy-conscious development habits among students early on. By exposing students to tools like \tool{} and its fine-grained profiling, educators highlight the significant energy demands of DL and the need for efficiency. Educators can prepare the next generation of DL developers to prioritize energy efficiency and build more sustainable AI systems by integrating \tool{} in class projects and assignments to measure and optimize model energy consumption, providing hands-on experience with Green AI principles.


\section{Conclusions and Future work}
In this work, we focused on the critical aspect of energy consumption in deep learning and introduced \tool{}, a fine-grained energy measurement framework.
\tool{} uses static instrumentation to segregate the execution of an \api{} and to ensure machine's stability.
Our experiments and evaluation have shown that the proposed framework measures consumed energy at the \api{} granularity.
Our empirical analysis investigating the influence of input parameter data size and execution time on energy consumption reveals that an \api{}'s energy consumption shows a linear relationship with input data size.
Furthermore, we consolidated and categorized various considerations, challenges, and issues we faced throughout the design and development of the framework. Addressing these challenges will guide future efforts in creating fine-grained energy measurement tools.
In the future, we would like to use the proposed framework to extend our empirical analysis to investigate additional aspects related to the energy profile of \dl{} framework \api{}s. Exploring the role of hyper-parameters and data quality on fine-grained energy consumption will further enhance energy profiling capabilities. Additionally, we plan to extend the application of \tool{} to other popular \dl{}frameworks like \pytorch{} for a comprehensive analysis of energy-efficient models.

\section{Acknowledgement}
Saurabhsingh Rajput and Tushar Sharma are supported by the Natural Sciences and Engineering Research Council of Canada (NSERC) through grant NSERC Discovery RGPIN/04903.
Maria Kechagia and Federica Sarro are supported by the European Research Council under grant no. 741278 (EPIC).

\bibliographystyle{ACM-Reference-Format}

\bibliography{references}
\end{document}